\documentclass{article}

\usepackage[final]{neurips_2024}

\usepackage[utf8]{inputenc} %
\usepackage[T1]{fontenc}    %
\usepackage{hyperref}       %
\usepackage{url}            %
\usepackage{booktabs}       %
\usepackage{amsfonts}       %
\usepackage{nicefrac}       %
\usepackage{microtype}      %
\usepackage{xcolor}         %

\usepackage{graphicx}
\usepackage{subcaption}
\usepackage{appendix}
\usepackage{multirow}
\usepackage{csquotes}
\usepackage{amsmath}

\definecolor{burgundy}{rgb}{0.7, 0.0, 0.13}

\title{Evaluating Synthetic Activations composed of SAE Latents in GPT-2}

\author{
  Giorgi Giglemiani $^*$ \\
  LASR Labs\\\
  \texttt{giglemiani@gmail.com}
  \And
  Nora Petrova $^*$ \\
  LASR Labs\\\
  \texttt{nora.axion@gmail.com}
  \And
  Chatrik Singh Mangat $^*$ \\
  LASR Labs\\\
  \texttt{chatrikmangat@outlook.com}
  \AND
  Jett Janiak \\
  LASR Labs\\\
  \texttt{jettjaniak@gmail.com}
  \And
  Stefan Heimersheim \\
  Apollo Research\\\
  \texttt{stefan@apolloresearch.ai}
}

\begin{document}

\maketitle
\def\thefootnote{*}\footnotetext{Equal contribution. Correspondence to \href{mailto:giglemiani@gmail.com}{giglemiani@gmail.com}}

\begin{abstract}
  Sparse Auto-Encoders (SAEs) are commonly employed in mechanistic interpretability to decompose the residual stream into monosemantic SAE latents. Recent work demonstrates that perturbing a model’s activations at an early layer results in a step-function-like change in the model's final layer activations. Furthermore, the model's sensitivity to this perturbation differs between model-generated (real) activations and random activations. In our study, we assess model sensitivity in order to compare real activations to synthetic activations composed of SAE latents. Our findings indicate that synthetic activations closely resemble real activations when we control for the sparsity and cosine similarity of the constituent SAE latents. This suggests that real activations cannot be explained by a simple \enquote{bag of SAE latents} lacking internal structure, and instead suggests that SAE latents possess significant geometric and statistical properties. Notably, we observe that our synthetic activations exhibit less pronounced activation plateaus compared to those typically surrounding real activations.
\end{abstract}

\section{Introduction} \label{sec:intro}

Neural networks often exhibit polysemanticity, where individual neurons fire for multiple features \citep{olah2017feature}. This complexity poses a significant challenge for interpreting computation in models, as it obscures the direct relationship between neuronal activations and specific semantic concepts. To explain polysemanticity, the theory of \emph{superposition} suggests that neural networks represent more features than they have dimensions. These features are linearly represented as directions in activation space which form an overcomplete basis \citep{elhage2022solu, bricken2023monosemanticity}. Although there is evidence for many linearly represented features, the claim that all features of a neural network are represented directions remains more speculative \citep{engels2024languagemodelfeatureslinear,smith2024superposition,olah2024multidimentional_features}.

Sparse Auto-Encoders (SAEs) decompose a model’s residual stream into a sparse set of features \citep{Sharkey_Braun_Millidge_2022}. SAEs have become increasingly popular as a method to represent model activations in terms of more monosemantic and interpretable latents \citep{bricken2023monosemanticity,cunningham2023sparseautoencodershighlyinterpretable,templeton2024scaling}. As the community increases its reliance on SAEs to interpret model behavior, it becomes more important to verify that decompositions generated by SAEs accurately capture abstractions that are used by the model.

Neural networks employing superposition to represent features must address the challenge of interference to maintain performance \citep{hänni2024mathematicalmodelscomputationsuperposition}. This necessitates an ability to accurately extract individual features while mitigating noise from \enquote{nearby} features in the representation space. \citet{Heimersheim_sensitive_directions} observed two key phenomena related to this: activation plateaus and directional sensitivity. Both are characterized by changes in the L2 distance of model activations at the final layer in response to early layer perturbations. Activation plateaus represent regions around an activation within which perturbations do not affect the model output significantly, indicating model robustness to small amounts of noise. Importantly, these activation plateaus are present around model-generated activations (real activations) but not around random points sampled from the distribution of model-generated activations (random activations). This suggests the presence of an error correction mechanism that the model might be using to deal with interference. Directional sensitivity refers to the variation in the model’s response to perturbations based on the direction of the noise being added \citep{Lindsey2024saenote,Heimersheim_sensitive_directions}. The model's sensitivity to a particular direction is characterized by the inverse of the perturbation magnitude required in that direction to induce a step-function-like change (blowup) in the model output.

In this paper, we focus on generating synthetic activations composed of SAE latents and testing whether they behave like real activations. We specifically test the \enquote{bag of SAE latents} hypothesis and discover that arbitrary combinations of SAE latents do not produce activations that resemble real activations. We find that controlling for the sparsity and the cosine similarity of SAE latents enables us to create synthetic activations that most closely resemble the sensitivity of real activations. We test whether these observations generalize to activation plateaus and find that synthetic activations don't behave like real activations.

Our contributions are:
\begin{enumerate}
    \item We find that the \enquote{bag of SAE latents} approach is not sufficient to produce synthetic activations that resemble model-generated (real) activations.
    \item We find that the sparsity of the top SAE latent, the relative latent activations, and the cosine similarity between  the active latents and the top latent play an important role in determining whether synthetic activations behave like real activations. 
    \item The performance of synthetic activations in the sensitivity experiment does not transfer to the activation plateau experiment that we conduct. We find that synthetic activations do not have activation plateaus around them like real activations do.
\end{enumerate}

\section{Background}

Our experiments are based on the setup described in \citet{Heimersheim_sensitive_directions}, wherein they perturbed model activations at an early layer and measured the effect it had on the L2 distance of late-layer activations. They investigated activation plateaus and sensitive directions in \texttt{GPT-2}, motivated by the error correction mechanism predicted by computation in superposition. They explored two key predictions: (1) model-generated activations should be resistant to small perturbations, exhibiting "activation plateaus", and (2) perturbations towards model-generated activations should affect model output more quickly than towards random directions. Their findings supported both of their predictions, providing evidence for an error correction mechanism used by the model to suppress small amounts of noise. This research aimed to better understand computation in superposition and to find dataset-independent evidence for model features, potentially connecting to SAE research.

\section{Related Work} \label{sec:related}

Previous works have studied the model's response to residual stream perturbations using different experiments. We discuss the works that are the most relevant to ours below.

\citet{janiak2024characterizingstableregionsresidual} identified stable regions (corresponding to activation plateaus) in the activation space of transformer-based models, hypothesizing their role in error correction and semantic distinctions. Our work primarily focuses on sensitive directions, though we study activation plateaus around synthetic activations and compare them against real activations.

\citet{Gurnee2024reconstruction_pathological} found that substituting the model activation at an early layer with its SAE reconstruction causes a bigger jump in KL divergence of the model’s next-token prediction probabilities than substituting it with a random vector that is the same L2 distance away from the original activation as the SAE reconstruction. While our work focuses on compositions of SAE latents, we study the effect of SAE reconstruction error on our experiments (Appendix \ref{app:recon}).

\citet{Lee2024sensitive_directions} found that  SAE reconstruction errors are not pathologically large when compared to more realistic baselines. They also found that end-to-end SAE latents do not exhibit stronger effects on model output compared to traditional SAE latents. Their work focuses on perturbations in individual SAE latent directions while we study compositions of SAE latents.

\citet{Lindsey2024saenote} found that ablating an SAE latent had a significantly larger effect on model performance than doubling the latent’s activation. Additionally, they found that dampening latent activations had almost as strong of an effect on the output distribution as latent ablation. In our study, we focus on composing synthetic activations and studying SAE latent properties.

\section{Method} \label{sec:method}

Since model-generated (real) activations exhibit distinct behaviors in both experimental settings used in \citet{Heimersheim_sensitive_directions}, we test whether synthetic activations composed of SAE latents do the same. We adapt these experimental settings to study relationships between SAE latents that are important for generating synthetic activations that are in-distribution for the model. We discuss the methodology for the directional sensitivity experiment in Section \ref{sub:setup}, and discuss the different types of activations we test in Sections \ref{sub:control} and \ref{sub:synth}. Finally, we discuss the activation plateau experiment in Section \ref{sub:plateau}.

\subsection{Perturbation Setup} \label{sub:setup}

For each perturbation, we only make changes to the activations at the last token position in the context at layer 1 (\texttt{blocks.1.hook\_resid\_pre}). We refer to the original unperturbed activation as the base activation, often denoted by A. We perturb the base activation A by slowly adding increments towards a direction $D$:
\[
\rm A_{\mathrm{pert}}(n) = \mathrm{A} + 0.5 \cdot \mathrm{n} \cdot \mathrm{D}
\]
where $n$ is the step number, going from $0$ to $100$, and $D$ is a unit vector. For all our perturbations, we define the direction $D$ as the normalised difference between a base activation and a target activation. We use a step size of $0.5$ because this makes the perturbation norm comparable to the typical norm of activations, which is $\simeq 56$.

For each perturbation, we compute the L2 distance between the activations of the original and the perturbed run after the final layer (\texttt{blocks.11.hook\_resid\_post}). We use this L2 distance to study the effect of our perturbations instead of the KL divergence of the next-token prediction probabilities. We focus on L2 distance because KL divergence plots tend to obscure the structure of activation plateaus (we still include KL divergence based results in Appendix \ref{app:kldiv} for comparison to previous work).

We test multiple metrics in order to formalize the location of the blowup in L2 distance, and we find that the maximum slope of the L2 distance against the perturbation step curve represents it most cleanly (shown in Figure \ref{fig1}; see Appendix \ref{app:metrics} for the other metrics we test). For each perturbation, we label the maximum slope (MS) step and use it to refer to the location of the blowup further in this work.

For all our experiments, we use \texttt{GPT2-small} \citep{radford2019language} and run inference on randomly sampled prompts of sequence length $10$ from the \texttt{OpenWebText} dataset\footnote{Using a tokenized version of this dataset available \href{https://huggingface.co/datasets/apollo-research/Skylion007-openwebtext-tokenizer-gpt2}{here}.} \citep{Gokaslan2019OpenWeb}, and we collect model-generated activations from the residual stream at Layer 1 (\texttt{blocks.1.hook\_resid\_pre}). We use \texttt{GPT2-small} SAEs \citep{Bloom2024sae}, \texttt{sae-lens} \citep{bloom2024saetrainingcodebase} and \texttt{TransformerLens} \citep{nanda2022transformerlens} to perform our experiments and generate synthetic activations. 

\subsection{Non-SAE Baselines} \label{sub:control}

\begin{figure}[h]
  \centering
  \begin{subfigure}{0.497\textwidth}
  \includegraphics[width=\textwidth]{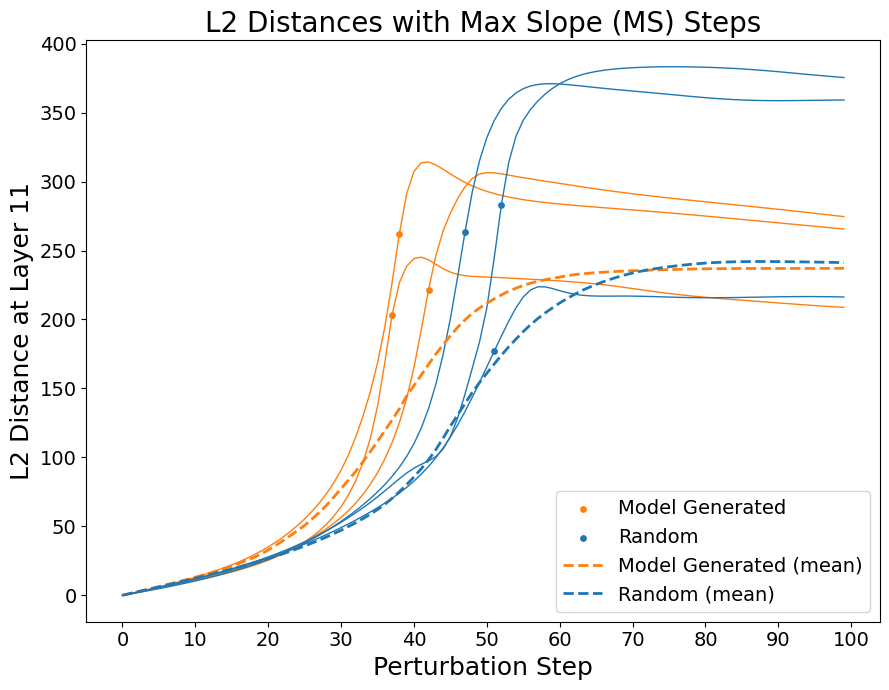}
  \end{subfigure}  
  \begin{subfigure}{0.497\textwidth}
  \includegraphics[width=\textwidth]{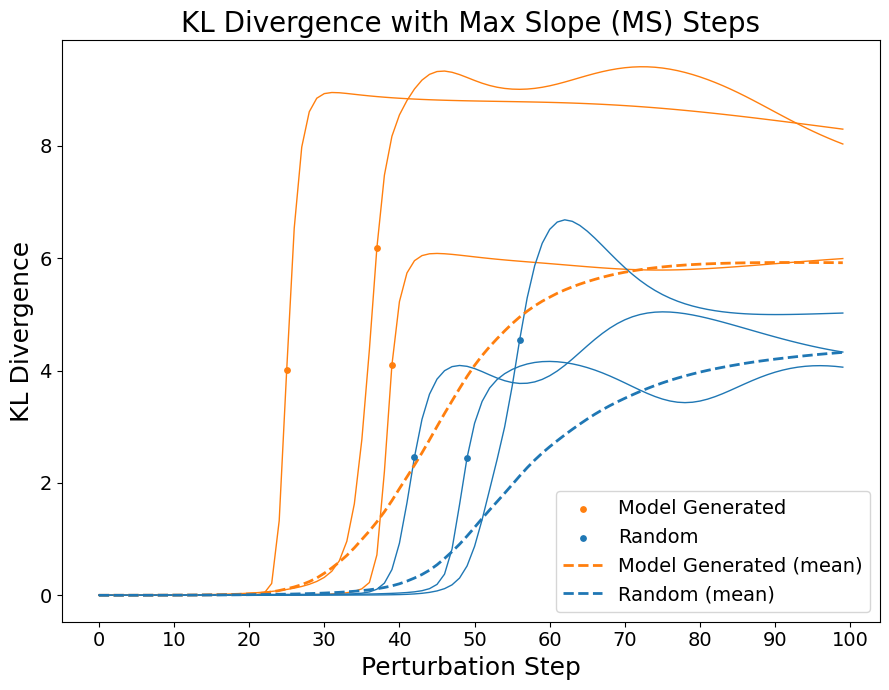}
  \end{subfigure}
  \caption{\footnotesize{The L2 distance after Layer 11 (left) and the KL divergence of the next-token prediction probabilities (right) between the perturbed and unperturbed model, as three base activations at Layer 1 are slowly perturbed towards model-generated activations (orange) and random points sampled from the distribution of model activations (blue). The x-axis represents the total length of the perturbation broken into $100$ steps of size $0.5$ each. The dot on each solid line represents the maximum slope (MS) step for each perturbation. The dashed lines represent the average L2 distance and KL divergence per step for $1000$ perturbations of both types. The linear part at the start of the curves represents the activation plateau, and the sharp rise in the curves represents the blowup.}}
  \label{fig1}
\end{figure}

In order to compare our setup to previous work \citep{Heimersheim_sensitive_directions}, we run perturbations towards model-generated (real) and random activations. We sample $1000$ prompts from the dataset and run inference on them to obtain the base activations, and perturb each base activation in two directions:
\begin{enumerate}
    \item \textbf{Model-generated (real):} Towards a randomly selected activation produced by the model.
    \item \textbf{Random:} Towards a randomly sampled point from a normal distribution with the same mean and covariance as model-generated activations (calculated using $32,000$ model-generated activations).
\end{enumerate}
We plot examples of perturbations towards real and random activations in Figure \ref{fig1}. The distribution of distances between the base activation and the target are similar for both baselines, with a mean of $\simeq 40$ in activation space. Additionally, we find that the average cosine similarity between two model-generated activations with respect to the SAE decoder bias is $\simeq 0.42$.

\subsection{Synthetic Activations} \label{sub:synth}

We construct synthetic activations using three methods that each use different amounts of information about SAE latents, and compare which ones match the behavior of real activations.

Our simplest approach (\textbf{synthetic-random}) randomly selects SAE latents and assigns them the same latent activations as the active latents of the base activation. We consider this a weak approach to compose synthetic activations and mainly focus on the next two methods, but we show the results for this method in Appendix \ref{app:relative}.

Our second approach (\textbf{synthetic-baseline}) accounts for SAE latent sparsities along with latent activations, matching the distribution of latent sparsities from the base activation:
\begin{enumerate}
\item We use the SAE to encode the base activation and obtain its active latents and their latent activations.
\item We replace each active SAE latent with a new one sampled from $10$ most similarly-sparse latents and assign it the latent activation of the original active latent.
\item We decode the new latent activations to obtain the \textbf{synthetic-baseline} activation.
\end{enumerate}

For our third and final method (\textbf{synthetic-structured}) we capture and reproduce geometric properties that SAE latents of real activations have. We get the best results when we control for sparsity of latents and the cosine similarity relationship between latents. The procedure for the generation of synthetic-structured activations is as follows:
\begin{enumerate}
    \item We use the SAE to encode the base activation and obtain its active latents and their latent activations. The active latent with the highest latent activation is the top latent of the base activation ($\texttt{top\_base}$).
    \item We create a list of $100$ non-dead SAE latents with the most similar sparsity to $\texttt{top\_base}$.
    \item Out of the $100$ selected latents, we select one latent that has cosine similarity closest to $0.42$ (mean cosine similarity between two real activations w.r.t. the SAE decoder bias) with $\texttt{top\_base}$. \item This latent becomes the top latent for our synthetic activation ($\texttt{top\_synth}$), and we give it a latent activation value equal to that of $\texttt{top\_base}$.
    \item For each remaining active latent ($\texttt{l\_base}$) in the base activation:
    \begin{enumerate}
        \item We calculate its cosine similarity ($\texttt{l\_top\_cos\_sim}$) with \texttt{top\_base}.
        \item We select a latent (\texttt{l\_synth}) that has cosine similarity with \texttt{top\_synth} equal to \texttt{l\_top\_cos\_sim}.
        \item We assign \texttt{l\_synth} a latent activation value equal to that of \texttt{l\_base}.
    \end{enumerate}
    \item We construct a latent activation vector with zeros for all latents except the latents selected above, and decode it to obtain the \textbf{synthetic-structured} activation.
\end{enumerate}
We perform $1000$ perturbations towards synthetic-baseline and synthetic-structured activations each, using the setup described in Section \ref{sub:setup}, where the direction $D$ is the normalised difference between the synthetic activation and the base activation.

\subsection{Activation Plateaus} \label{sub:plateau}

If our synthetic activations behave like real activations, they should not only reproduce directional sensitivity behavior, but also exhibit activation plateaus. To test this, we use the following perturbation approach:
\begin{enumerate}
    \item We initiate perturbations from four distinct base activations: model-generated activations, synthetic activations generated using SAE latents of similar sparsity to the base activation (synthetic-baseline), synthetic activations generated using selected SAE latents based on sparsity and cosine similarity relationship (synthetic-structured), and random points sampled from the distribution described in Section \ref{sub:control} (random). 
    \item For all four starting points, we perturb towards a random activation.
    \item We track how quickly L2 distance at Layer 11 increases near the start of the perturbation by recording the activation plateau perturbation step (AP step) at which the L2 distance between the unperturbed and perturbed models crosses a value of $20$ (we found that a simple threshold was enough to distinguish the behavior of different activations).
    \end{enumerate}
The AP step measures the flatness of the activation plateau around activations, with larger AP steps signify flatter activation plateaus. We repeat this process for $1000$ perturbations and collect the distributions of AP steps for each activation type.

\section{Results} \label{sec:results}

We find that the behavior of synthetic activations as compared to random activations and real activations differs for the two experiments we perform. This suggests that directional sensitivity and activation plateaus point to different properties of SAE latents that make up real activations (see Appendix \ref{app:stats} for details on properties that we account for). We now provide detailed results for each of our experiments below.

\subsection{Directional Sensitivity} \label{sub:sensitivity_res}

\begin{figure}[h]
  \centering
  \begin{subfigure}{0.497\textwidth}
  \includegraphics[width=\textwidth]{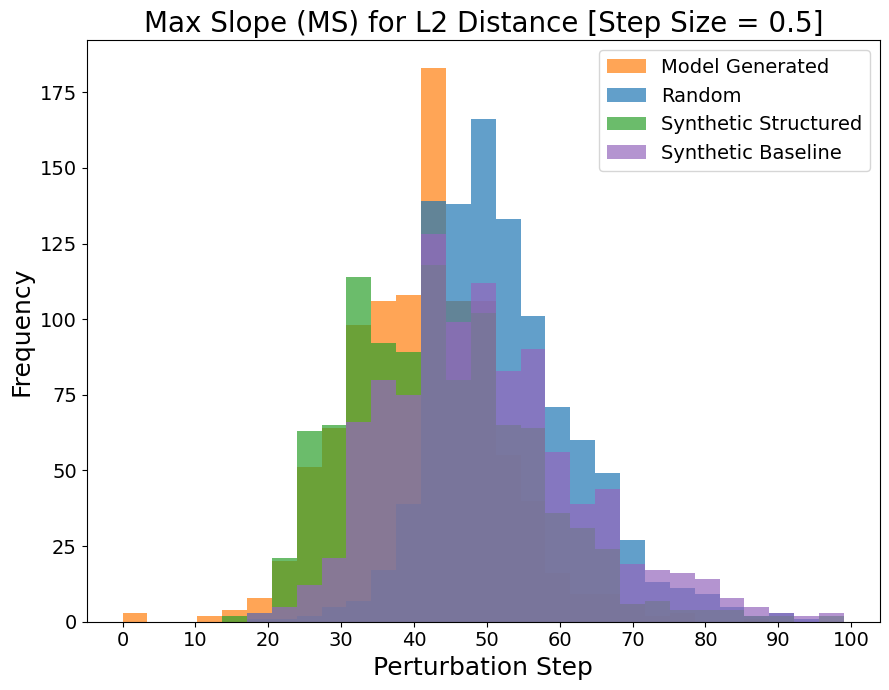}
  \end{subfigure}  
  \begin{subfigure}{0.497\textwidth}
  \includegraphics[width=\textwidth]{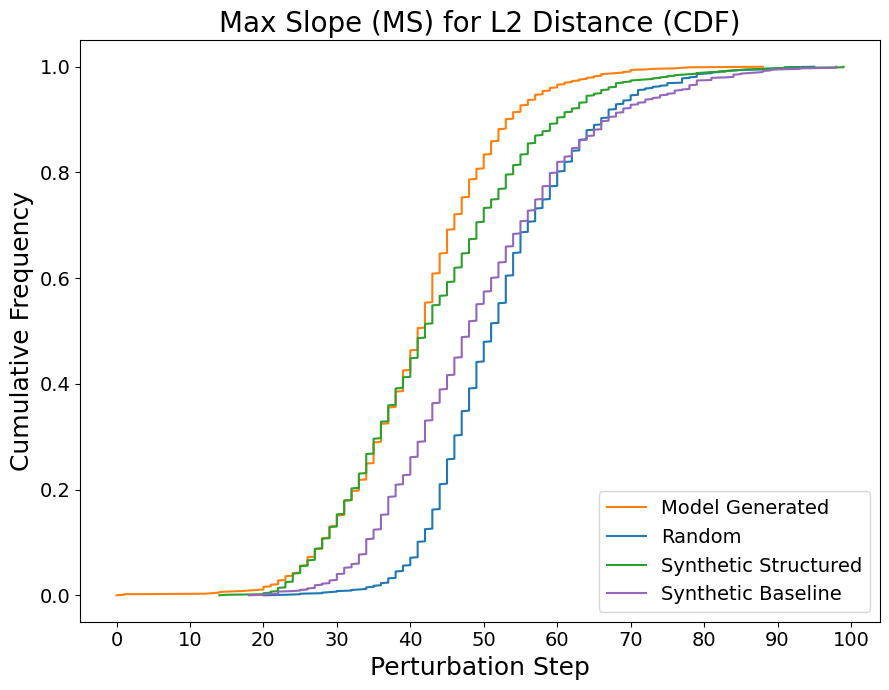}
  \end{subfigure}
  \caption{\footnotesize{The distributions of the max slope (MS) steps for perturbations towards model-generated (orange), random (blue), synthetic-baseline (purple), and synthetic-structured (green) activations. The left panel shows the counts of MS steps occurring in different bins along the length of the perturbation, and the right panel shows corresponding cumulative frequency. We find that perturbing towards synthetic-structured activations is more similar to perturbing towards model-generated activations as compared to perturbing towards synthetic-baseline activations.}}
  \label{fig2}
\end{figure}

For this experiment, we perturb real activations towards different types of activations we construct and study the model's sensitivity to these perturbations. Figure \ref{fig2} shows the distributions of max slope (MS) steps for L2 distance for perturbations towards synthetic-baseline and synthetic-structured activations compared to model-generated (real) and random activations. We also provide statistics for the MS step distributions for perturbations towards all activations types in Table \ref{tab1}.

When comparing the MS steps (Table \ref{tab1}) for perturbations towards real and random activations, we find that perturbations towards real activations cause earlier (lower mean) and more localized (lower variance) blowups than perturbations towards random activations. This means that perturbing towards real activations affects the model output more than perturbing towards random activations.

While synthetic-baseline activations do not fully replicate the behavior of real activations, they still perform better than random activations (Figure \ref{fig2}). We calculate similarity between two MS step distributions using the Kolmogorov Smirnov (KS) statistic \citep{10.1214/aoms/1177730256}. This demonstrates that the model is more sensitive to perturbations towards synthetic activations composed of SAE latents than perturbations towards randomly sampled points from the distribution of model activations (random). This suggests that SAE latents encode more information about model computation than random directions do.

Importantly, perturbations towards synthetic-structured activations look a lot more similar to perturbations towards model-generated activations than perturbations towards synthetic-baseline and random activations do (Figure \ref{fig2}, Table \ref{tab1}). This implies that relationships between SAE latents are important, and that model-generated activations are not approximated well by \enquote{bags of SAE latents}. Additionally, we find that synthetic-random activations perform worse than synthetic-baseline activations, confirming our decision to use the latter as a stronger baseline (see Appendix \ref{app:relative} for more details on synthetic-random activations). 

The distance between the base activation and the target activation varies for each perturbation we perform, and we find that this can directly influence the location of the blowup in our setup. In order to remove the effect of the distance on the MS step distribution, we also perform perturbations with relative step size (see Appendix \ref{app:relative} for details). The gap between the synthetic-structured and synthetic-baseline reduces in the relative step size setup. This is because synthetic-baseline activations are further away from base activations than synthetic-structured activations, and hence cause blowups later.

\begin{table}[h]
  \centering
  \begin{tabular}{lcccc}
    \multicolumn{4}{c}{Max Slope (MS) step distribution statistics} \\
    \toprule
    Activation Type & Mean & Std dev & KS \\
    \midrule
    Model Generated & 41.11 & 10.40 & 0.00 \\
    Random & 52.49 & \textbf{10.21} & 0.45 \\
    Synthetic Baseline & 49.61 & 13.25 & 0.28 \\
    Synthetic Structured & \textbf{43.48} & 12.79 & \textbf{0.11} \\    
    \bottomrule
    \\
  \end{tabular}
  \caption{\footnotesize{When comparing perturbations towards different activations, we find that synthetic-structured activations behave more similar to model-generated activations than synthetic-baseline and random activations do. This table contains the mean, standard deviation and KS statistic for MS step distributions for all the perturbations we perform with fixed step size. The KS statistic is measured against perturbations towards model-generated activations, with lower values indicating higher similarity.}}
  \label{tab1}
\end{table}

\subsection{Activation Plateaus} \label{sub:plateau_res}

\begin{figure}[h]
  \centering
  \begin{subfigure}{0.497\textwidth}
  \includegraphics[width=\textwidth]{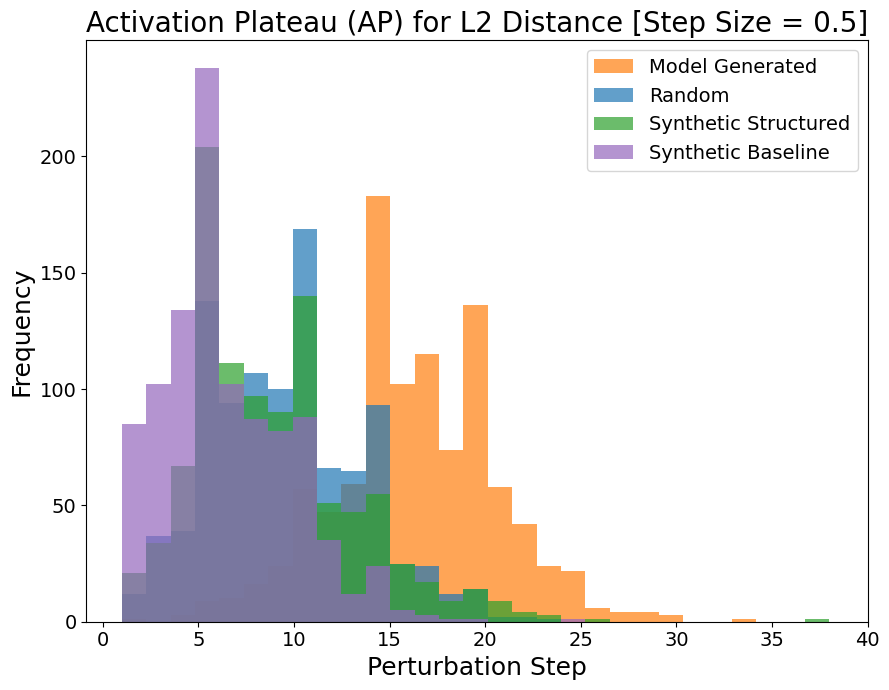}
  \end{subfigure}  
  \begin{subfigure}{0.497\textwidth}
  \includegraphics[width=\textwidth]{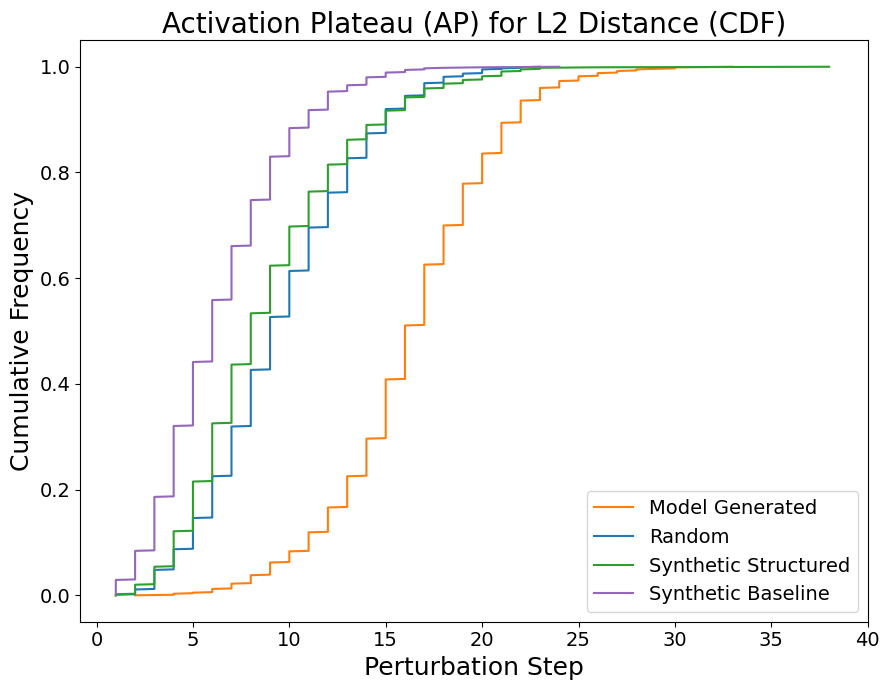}
  \end{subfigure}
  \caption{\footnotesize{The distributions of the activation plateau (AP) steps for perturbations starting at model-generated, random, synthetic-baseline, and synthetic-structured activations. We perturb towards random activations in all cases. The left panel shows the counts of AP steps occurring in different bins along the length of the perturbation, and the right panel shows the cumulative frequency for the same. We find that model-generated activations (orange) have flatter plateaus around them than all of the other activation types.  We also see that synthetic-baseline activations (purple) have the steepest plateaus around them, while plateaus around synthetic-structured (green) and random (blue) activations look similar.}}
  \label{fig3}
\end{figure}

For this experiment, we perturb different types of activations towards random directions to assess whether they have activation plateaus around them. Figure \ref{fig3} shows the distributions of AP steps for L2 distance for perturbations starting at synthetic-baseline and synthetic-structured activations compared to model-generated and random activations.

Our findings reveal varying sizes of plateaus around different types of activations. Model-generated activations exhibit more pronounced plateaus, indicating greater robustness to noise compared to synthetic and random activations, which display less distinct plateau regions. The relatively less pronounced plateaus around synthetic-baseline activations provide additional evidence against the \enquote{bag of SAE latents} approach, as this behavior notably differs from that of model-generated activations. While synthetic-structured activations show improvement, the differences in plateau characteristics suggests that they may not fully capture all significant relationships between SAE latents. To quantify the impact of SAE reconstruction error on the discrepancy between synthetic and model-generated activations, we conducted tests detailed in Appendix \ref{app:recon}, which indicate that this contribution is minimal.

\section{Limitations} \label{sec:discussion}

The heuristics we use to construct synthetic activations leave room for improvement, as evidenced by the gap between them and model-generated activations, especially for activation plateaus. We use cosine similarity between SAE latents to capture geometric relationships between them, but leave accounting for latent co-occurrence and other relationships between latents for future work.

We find that the synthetic activations we construct do not match the cosine similarity distribution of SAE latents in model-generated activations closely (see Appendix \ref{app:stats}). Our method also leverages information from the base activation in order to construct synthetic activations. While this is not ideal, we have verified that using information from a different model-generated activation for the construction does not change our results.

The L2 distance curves across perturbation steps exhibit significant variability, potentially impacting the effectiveness of our MS metric in identifying key steps. This metric assumes curve smoothness, which may not hold in practice (Figure \ref{fig1}). More robust metrics could potentially produce clearer results (see Appendix \ref{app:metrics} for our exploration of other metrics).

Our study focuses solely on one layer of \texttt{GPT2-small}, necessitating further investigation across different layers, models, and SAEs to establish broader applicability. Additionally, we confined our perturbations to the final token position only. Exploring different context lengths is crucial to assess the generalizability of our findings. These extensions would provide a more comprehensive understanding of the observed phenomena.

\section{Conclusion} \label{sec:conclusion}

We find additional evidence that \texttt{GPT-2} is more sensitive to perturbations towards model-generated activations than random directions, and that model-generated activations cannot simply be explained by \enquote{bags of SAE latents}. We find that leveraging statistical and geometric properties of SAE latents helps us create synthetic-structured activations which are more similar to model-generated activations. Yet, these synthetic-structured activations lack the characteristic plateaus surrounding model-generated activations, suggesting there may be additional SAE latent properties influencing model computation. 

This presents exciting avenues for future work on model sensitivity to perturbations:
\begin{itemize}
\item Coming up with better approaches to constructing synthetic activations that perform better in both experimental settings.
\item Checking for the existence of thresholds below which the model output does not respond to changes in latent activations.
\item Case studies which look at model sensitivity to perturbations in directions created using interpretable SAE latents and take context into account.
\item Zooming into latent ablation based perturbations further to study which latents contribute the most to blowups.
\end{itemize}

\section{Acknowledgements} \label{acknowledgements}

Erin Robertson, Charlie Griffin and the whole LASR Labs team for making this research project possible; Joseph Bloom for comments and SAEs used for this work; Daniel Lee for helpful discussions and for pointing us towards using NL as a metric; Jake Mendel for fundamental discussions about this research direction; Lawrence Chan, Nicholas Goldowsky-Dill, Bilal Chughtai, Daniel Tan, David Krueger, Joe Needham, David Chanin, Tomáš Dulka and Diogo Cruz for review and comments.

\bibliography{references}

\appendix

\section{Analyzing different perturbations setups and synthetic activations} \label{app:relative}
\setcounter{figure}{0}
\setcounter{table}{0}
\renewcommand\thefigure{\Alph{section}.\arabic{figure}}
\renewcommand\thetable{\Alph{section}.\arabic{table}}

In the main paper we use absolute step size for perturbations, however blowup locations have a dependence on the distance between the base and target activations, which can make the MS step distributions with absolute step size misleading. We know that the blowup location does not solely depend on the distance between the base and target activations, and in order to isolate this property, we create a distance agnostic setup using relative step sizes. In the relative step size approach, our perturbations always start at a base activation (A) and end at a target activation (T) using linear interpolation:
\[
\rm A_{\rm pert}(\rm n) = \left(1 - \frac{n}{100}\right) \cdot A + \frac{n}{100} \cdot T
\]
where $n$ is the perturbation step, which goes from $0$ to $100$. This method ensures that we always transition from the base activation to the target activation in a fixed number of steps, regardless of the distance between them. By using relative step size, we remove the dependence of the blowup location on distance, and instead compare the effect of perturbations purely in terms of the percentage of base and target activations present at each step. For example, step $50$ in this setup implies that the perturbed activation is made up of $50\%$ base activation and $50\%$ target activation. 

\begin{figure}[h]
  \centering
  \begin{subfigure}{0.497\textwidth}
  \includegraphics[width=\textwidth]{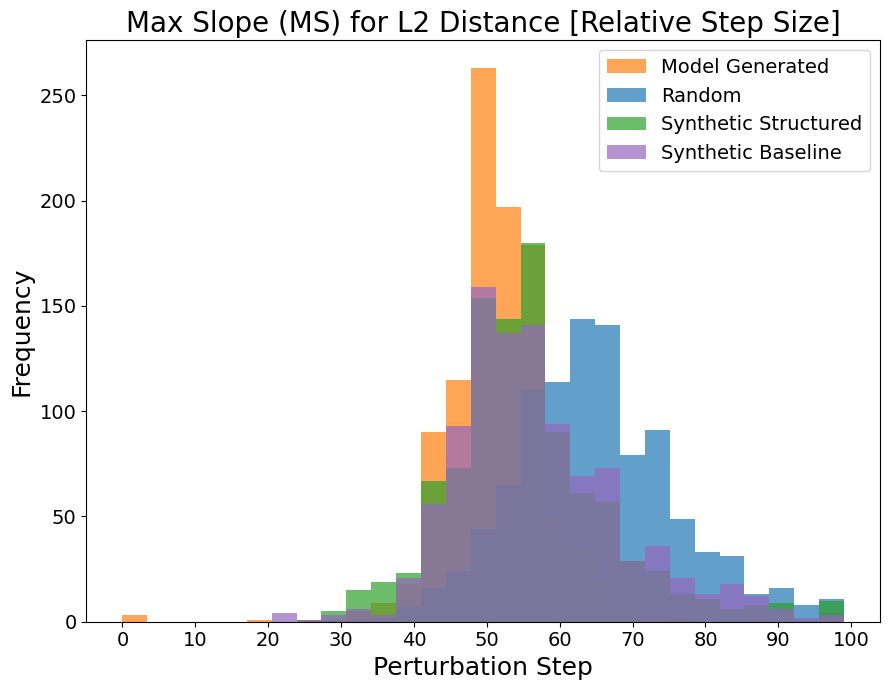}
  \end{subfigure}  
  \begin{subfigure}{0.497\textwidth}
  \includegraphics[width=\textwidth]{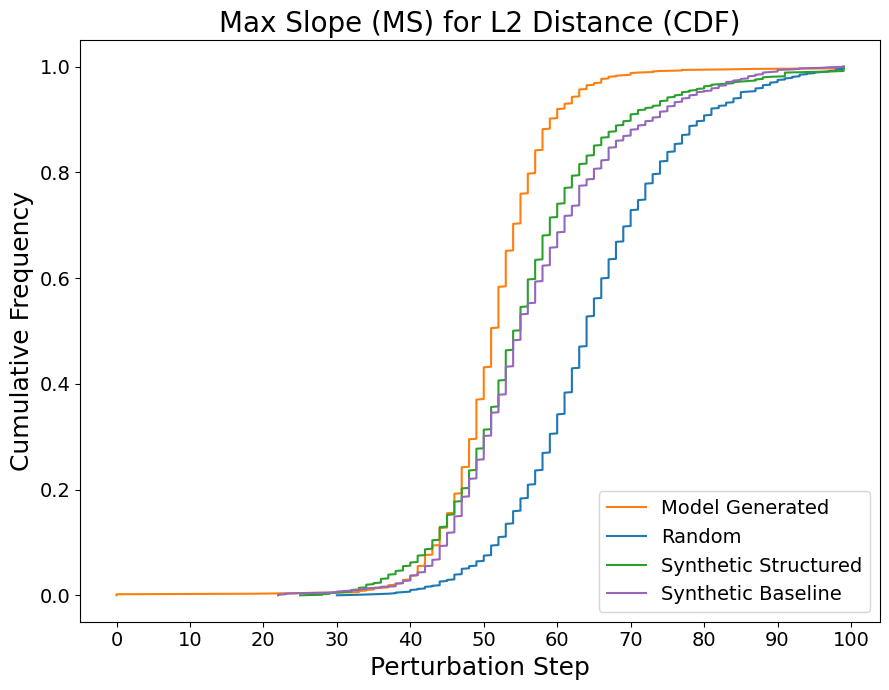}
  \end{subfigure}
  \caption{\footnotesize{The distributions of the max slope (MS) steps for perturbations with relative step size towards model-generated (orange), random (blue), synthetic-baseline (purple), and synthetic-structured (green) activations. The left panel shows the counts of MS steps occurring in different bins along the length of the perturbation, and the right panel shows the cumulative frequency for the same. We find that perturbing towards synthetic-structured activations in the relative step size setup is slightly more similar to perturbing towards model-generated activations than perturbing towards synthetic-baseline activations is.}}
  \label{figA1}
\end{figure}

In the relative step size setup, we find that the MS step distribution for perturbations towards model-generated activations peaks more strongly around step $50$ than in the absolute step size setup. The blowups are also localized between step $30$ and $70$, implying that blowups usually happen in the middle of the perturbation  (Figure \ref{figA1}). We posit that until step $30$, the model treats the interpolated activation as the base activation. This is due to $70\%$ of the interpolated activation coming from the base activation, and the remaining $30\%$ coming from the target activation being treated as noise. This effect reverses at step $70$, where the model starts treating the interpolated activation as the target activation, and the $30\%$ that comes from the base activation is considered noise. 

Our analysis reveals that the MS step distribution for random activation perturbations exhibits marginally higher variance than the absolute step size setup, with a rightward shift relative to the distribution for model-generated activation perturbations (Table \ref{tabA1}). This suggests that stronger perturbations towards random activations are required to induce a blowup compared to model-generated activations. Furthermore, it indicates that the model is more resilient to random noise than to noise directed towards another model-generated activation, requiring a greater magnitude of the former to cause confusion in the model.

In this setup, comparing perturbations with synthetic-baseline and synthetic-structured activations reveals that while synthetic-structured activations still more closely mimic model-generated activations, the disparity between the two has notably decreased (Figure \ref{figA1}, Table \ref{tabA1}). This suggests that synthetic-baseline activations less effectively align with the residual stream geometry of model-generated activations compared to synthetic-structured ones, explaining the latter's superior performance in the absolute step size setup.
Our findings indicate that considering latent sparsity is important for synthetic activations to emulate model-generated activations in the relative step size setup. Consequently, both synthetic-structured and synthetic-baseline outperform synthetic activations created using the \enquote{bag of SAE latents} approach without accounting for sparsity (synthetic-random).

We find that when we construct synthetic-structured activations (Section \ref{sub:synth}), omitting the cosine similarity constraint on the top latent and instead selecting based on sparsity similarity to the base activation's top latent yields the best-performing synthetic activations in the relative step size setup. However, these activations typically have greater distance from the base activation compared to synthetic-structured activations. Consequently, their performance in the absolute step size scenario is inferior to that of synthetic-structured activations.

\begin{table}[h]
  \centering
  \begin{tabular}{lccccccc}
    \multicolumn{7}{c}{Max Slope (MS) Step Distribution Statistics} \\
    \toprule
    \multirow{2}{*}{Activation Type} & \multicolumn{3}{c}{Absolute Step Size} & \multicolumn{3}{c}{Relative Step Size} \\
    \cmidrule(lr){2-4} \cmidrule(lr){5-7}
    & Mean & Std dev & KS & Mean & Std dev & KS \\
    \midrule
    Model Generated & 41.21 & 10.32 & 0.00 & 51.65 & 7.42 & 0.00 \\
    Random & 52.49 & 10.34 & 0.44 & 64.89 & 10.50 & 0.62 \\
    Synthetic Baseline & 49.88 & 12.60 & 0.31 & 57.07 & 11.29 & 0.27 \\
    Synthetic Structured & \textbf{43.45} & 12.78 & \textbf{0.11} & 55.69 & 11.31 & 0.22 \\
    Synthetic Random & 51.30 & \textbf{10.25} & 0.39 & 55.25 & \textbf{8.74} & 0.19 \\
    Synthetic Structured (w/o cos sim) & 50.17 & 11.96 & 0.31 & \textbf{54.47} & 10.68 & \textbf{0.17} \\
    \bottomrule
    \\
  \end{tabular}
  \caption{\footnotesize{We find that controlling for the sparsity of the top latent and the cosine similarity between the active latents play an important role in making synthetic-structured activations perform well in both absolute and relative step setups. This table contains the mean, standard deviation and KS statistic for MS step distributions for all types of synthetic activations we tested. The KS statistic is measured against perturbations towards model-generated activations, with a lower value meaning higher similarity. The entries in bold show the best match with statistics for model-generated activations.}}
  \label{tabA1}
\end{table}

\newpage
\section{Metrics for analysing blowups} \label{app:metrics}
\setcounter{figure}{0}
\setcounter{table}{0}
\renewcommand\thefigure{\Alph{section}.\arabic{figure}}
\renewcommand\thetable{\Alph{section}.\arabic{table}}

In our main analysis, we focus on the maximum slope (MS) as an indicator of the blowup step. In this section we share findings using the Area Under the Curve (AUC) and Non Linear (NL) metrics to represent important parts of the L2 distance vs perturbation step curve.

\subsection{Area Under Curve (AUC)} \label{sub:auc}

Our experimental results reveal that certain L2 distance curves deviate from the expected step-function-like pattern, causing the MS step to misrepresent the actual blowup location for these curves. In contrast, the AUC metric provides a more comprehensive assessment of activation behavior across the entire perturbation process. This approach not only identifies the steepest increase point but also effectively screens out atypical curves that might otherwise evade detection. AUC calculates the step at which the following ratio is maximized: 
\[
R = \textrm{area of the triangle defined by (0,0), (x,0) and (f(x),x)} / \textrm{area under the curve f(x)}
\]
where $f(x)$ is L2 distance as a function of the perturbation step $x$. This method is sensitive to the concavity or convexity of the perturbation curve. For predominantly concave curves (where the rate of change increases over time), the AUC blowup step tends to occur later, as the triangular area takes longer to outpace the actual area under the curve. Conversely, for convex curves (where the rate of change decreases over time), the AUC blowup step tends to occur earlier. This property allows the AUC method to implicitly capture information about the shape of the perturbation.

\begin{figure}[h]
  \centering
  \begin{subfigure}{0.497\textwidth}
  \includegraphics[width=\textwidth]{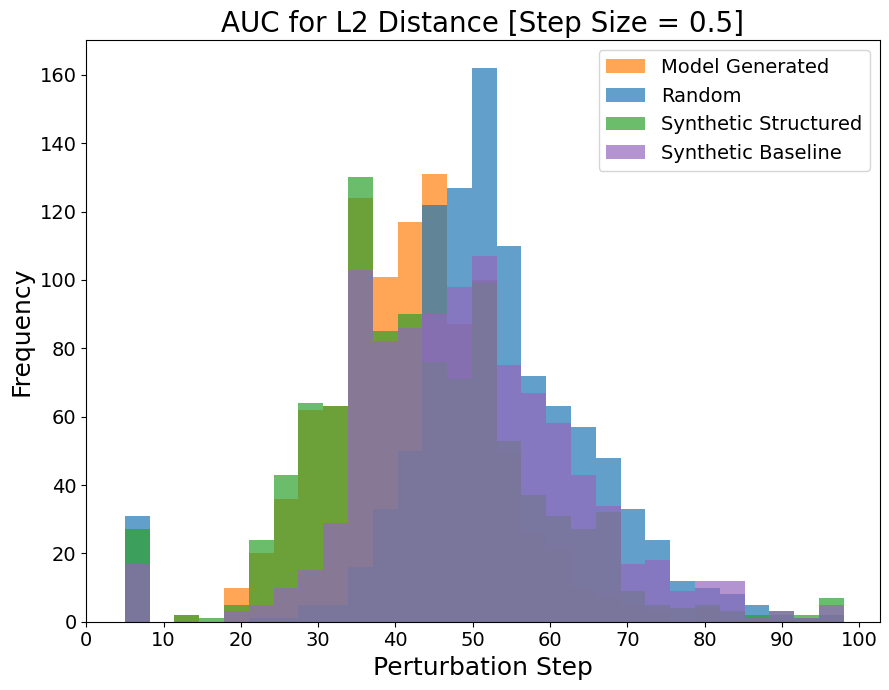}
  \end{subfigure}  
  \begin{subfigure}{0.497\textwidth}
  \includegraphics[width=\textwidth]{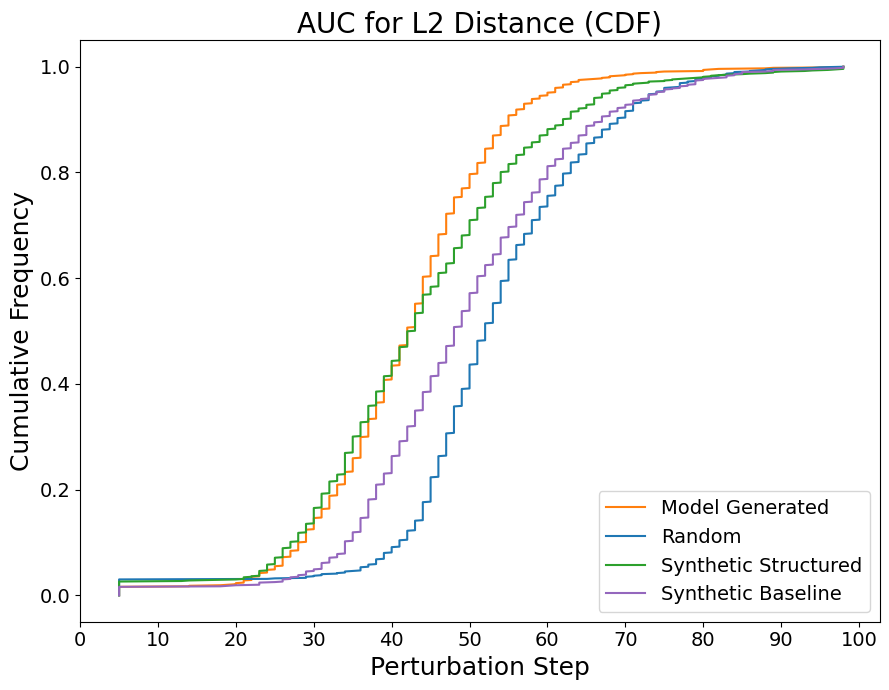}
  \end{subfigure}
  \begin{subfigure}{0.497\textwidth}
  \includegraphics[width=\textwidth]{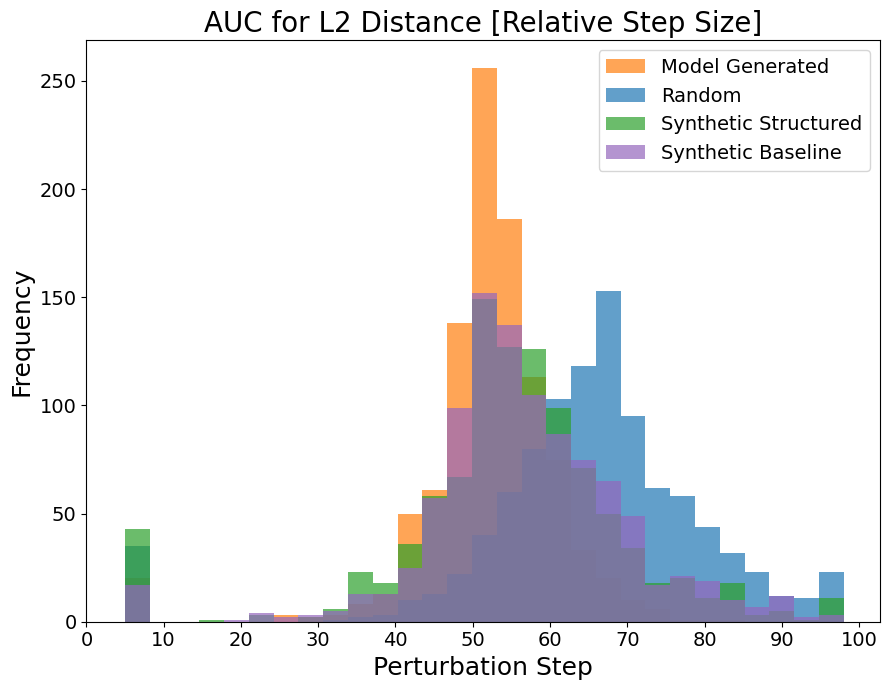}
  \end{subfigure}  
  \begin{subfigure}{0.497\textwidth}
  \includegraphics[width=\textwidth]{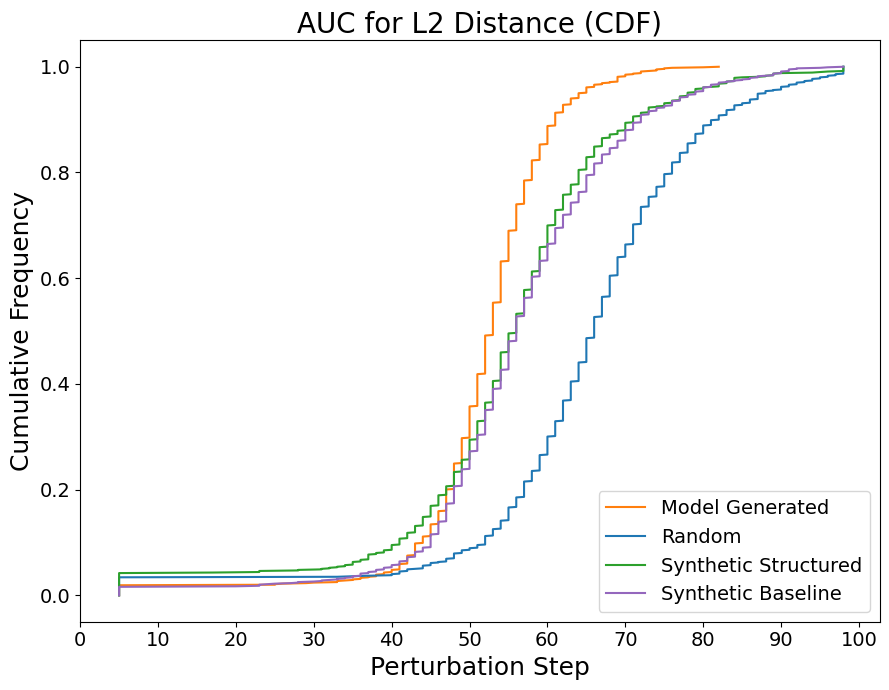}
  \end{subfigure}
  \caption{\footnotesize{The distributions of the AUC steps for perturbations with absolute step size (top) and relative step size (bottom) towards model-generated (orange), random (blue), synthetic-baseline (purple), and synthetic-structured (green) activations. The left column shows the counts of AUC steps occurring in different bins along the length of the perturbation, and the right column shows the cumulative frequency for the same. We find that our results for the AUC step distributions are similar to those for the MS step distributions.}}
  \label{figB1}
\end{figure}

The AUC metric serves as sanity check, confirming that most perturbations align with expectations. Convex L2 distance curves yield early AUC peaks, and Figure \ref{figB1} demonstrates that the majority of perturbations exhibit the anticipated concave shape. We find that our perturbation results hold for AUC step distributions in the absolute step size setup (Table \ref{tabB1}), with structured-synthetic activations more closely mimicking model-generated activations compared to synthetic-baseline activations. In the relative step size setup (detailed in Appendix \ref{app:relative}), synthetic-structured and synthetic-baseline activations perform similarly. This can be attributed to the higher prevalence of convex curves in perturbations towards synthetic-structured activations versus synthetic-baseline activations.

\begin{table}[h]
  \centering
  \begin{tabular}{lccccccc}
    \multicolumn{7}{c}{Area Under Curve (AUC) Step Distribution Statistics} \\
    \toprule
    \multirow{2}{*}{Activation Type} & \multicolumn{3}{c}{Absolute Step Size} & \multicolumn{3}{c}{Relative Step Size} \\
    \cmidrule(lr){2-4} \cmidrule(lr){5-7}
    & Mean & Std dev & KS & Mean & Std dev & KS \\
    \midrule
    Model Generated & 41.94 & 11.78 & 0.00 & 51.98 & 9.64 & 0.00 \\
    Random & 52.73 & 13.66 & 0.43 & 64.97 & 16.01 & 0.59 \\
    Synthetic Baseline & 49.31 & 14.20 & 0.25 & 56.66 & 13.20 & 0.22 \\
    Synthetic Structured & 43.54 & 14.99 & 0.09 & 54.84 & 15.51 & 0.21 \\
    \bottomrule
    \\
  \end{tabular}
  \caption{\footnotesize{We find that our results for the AUC step distributions are similar to those for the MS step distributions. This table contains the mean, standard deviation and KS statistic for AUC step distributions for all the perturbations we perform. The KS statistic is measured against perturbations towards model-generated activations, with a lower value meaning higher similarity.}}
  \label{tabB1}
\end{table}

\newpage
\subsection{Non-Linear (NL)} \label{sub:nl}

Using L2 distance to observe the perturbations reveals that the region before the blowup is not flat, but linear with varying slopes (Figure \ref{fig1}). In order to study the size of the initial linear portion of the curves, we use the Non-Linear (NL) metric, which points to the earliest step at which the slope of the L2 distance vs perturbation step curve deviates from linearity by more than $10\%$ of the initial slope. We use this metric as an alternate measure for the size of the activation plateau around the base activation along different perturbation directions.

\begin{figure}[h]
  \centering
  \begin{subfigure}{0.497\textwidth}
  \includegraphics[width=\textwidth]{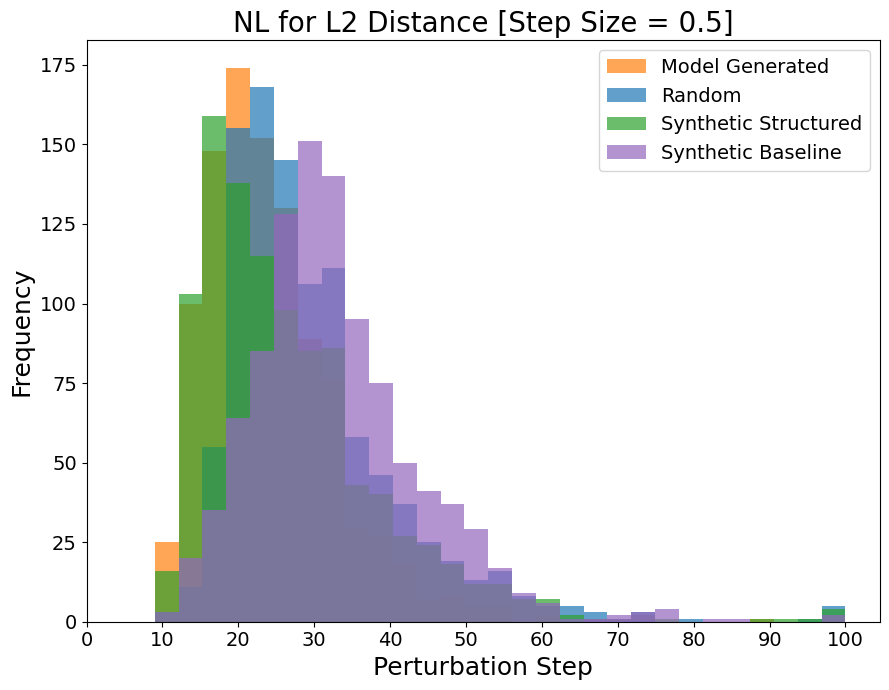}
  \end{subfigure}  
  \begin{subfigure}{0.497\textwidth}
  \includegraphics[width=\textwidth]{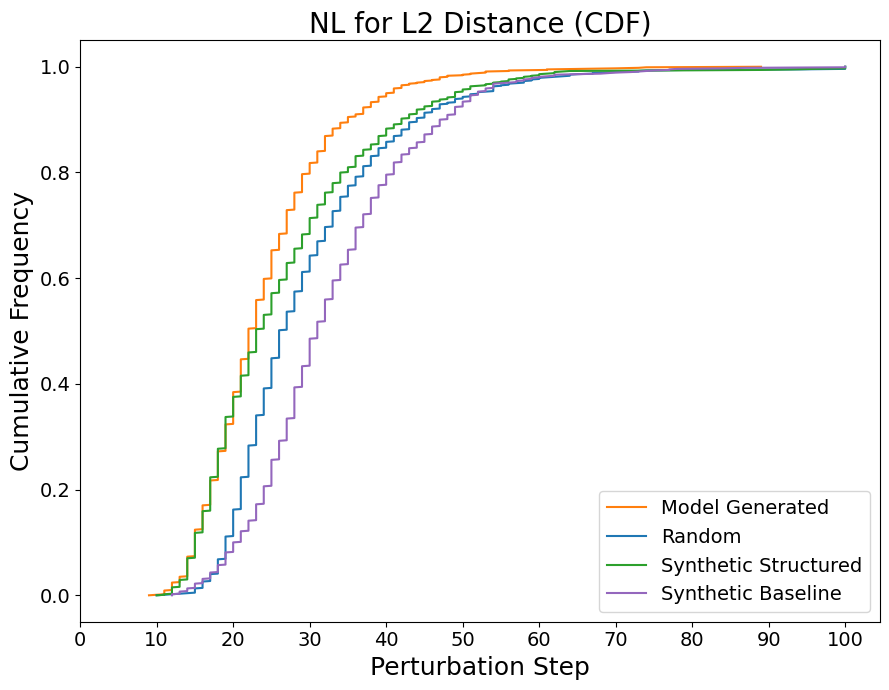}
  \end{subfigure}
  \begin{subfigure}{0.497\textwidth}
  \includegraphics[width=\textwidth]{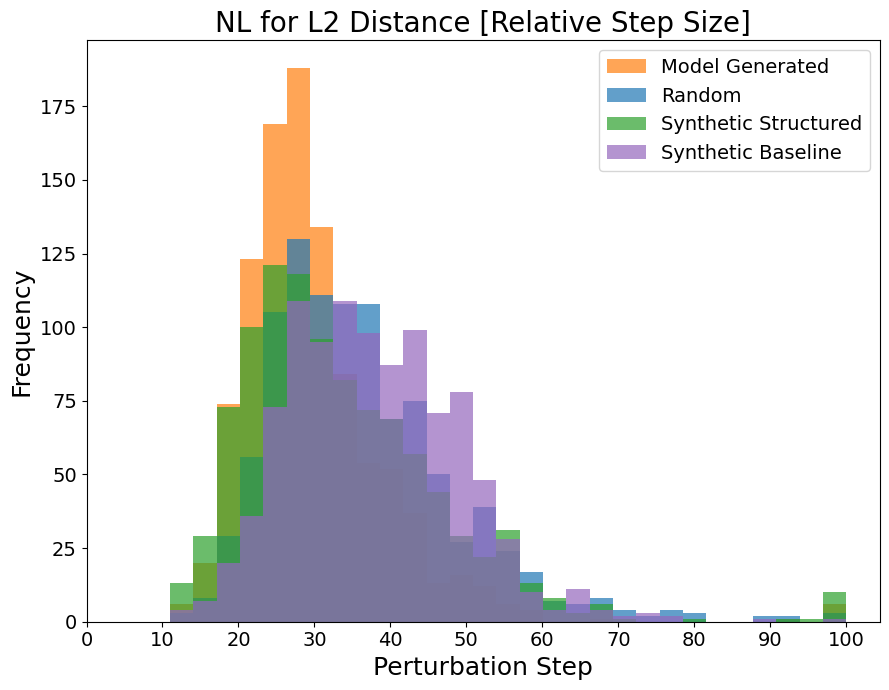}
  \end{subfigure}  
  \begin{subfigure}{0.497\textwidth}
  \includegraphics[width=\textwidth]{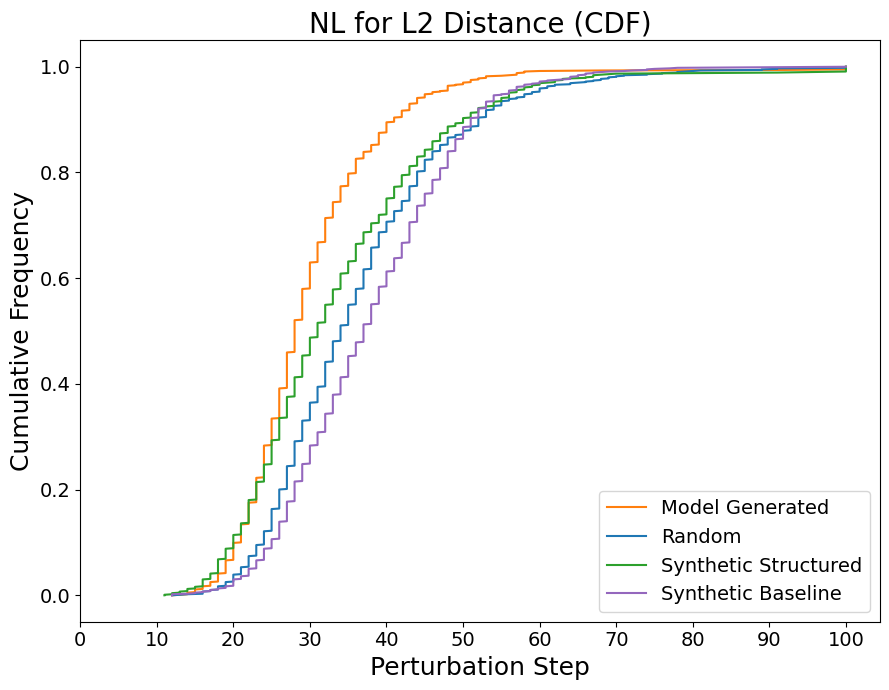}
  \end{subfigure}
  \caption{\footnotesize{The distributions of the NL steps for perturbations with absolute step size (top) and relative step size (bottom) towards model-generated (orange), random (blue), synthetic-baseline (purple), and synthetic-structured (green) activations. The left column shows the counts of NL steps occurring in different bins along the length of the perturbation, and the right column shows the cumulative frequency for the same. We find that synthetic-structured activations and random activations behave more like model-generated activations than synthetic-baseline activations do.}}
  \label{figB2}
\end{figure}

We observe that perturbations towards model-generated activations cause the quickest deviation from linearity followed by synthetic-structured activations, which is in line with our previous results for blowup locations (Figure \ref{figB2}, Table \ref{tabB2}). However, we find that the deviation from linearity occurs the latest during perturbations towards synthetic-baseline activations, which suggests that L2 distance has a higher initial slope for these perturbations, giving more room for changes in the slope before they are classified as a deviation from linearity. In this case, the behavior of synthetic-baseline activations provides further evidence that local relationships between SAE latents are important to approximate model-generated activations.

\begin{table}[h]
  \centering
  \begin{tabular}{lccccccc}
    \multicolumn{7}{c}{Non-Linear (NL) Step Distribution Statistics} \\
    \toprule
    \multirow{2}{*}{Activation Type} & \multicolumn{3}{c}{Absolute Step Size} & \multicolumn{3}{c}{Relative Step Size} \\
    \cmidrule(lr){2-4} \cmidrule(lr){5-7}
    & Mean & Std dev & KS & Mean & Std dev & KS \\
    \midrule
    Model Generated & 24.17 & 8.87 & 0.00 & 29.98 & 9.95 & 0.00 \\
    Random & 29.80 & 11.72 & 0.22 & 36.33 & 12.33 & 0.27 \\
    Synthetic Baseline & 32.90 & 11.25 & 0.40 & 37.91 & 10.96 & 0.37 \\
    Synthetic Structured & 26.72 & 12.25 & 0.11 & 33.69 & 13.44 & 0.17 \\
    \bottomrule
    \\
  \end{tabular}
  \caption{\footnotesize{In terms of NL step distributions, we find that synthetic-structured activations perform better than random activations, but synthetic-baseline activations do not. This table contains the mean, standard deviation and KS statistic for NL step distributions for all the perturbations we perform. The KS statistic is measured against perturbations towards model-generated activations, with a lower value meaning higher similarity.}}
  \label{tabB2}
\end{table}

\newpage
\section{KL Divergence} \label{app:kldiv}
\setcounter{figure}{0}
\setcounter{table}{0}
\renewcommand\thefigure{\Alph{section}.\arabic{figure}}
\renewcommand\thetable{\Alph{section}.\arabic{table}}

While previous works have predominantly used KL divergence as a measure of sensitivity, our analysis revealed potential limitations of this approach. We observed that KL divergence produces a step-function-like curve even when linear perturbations are performed at the final layer of the model right before the unembedding. This behavior suggests that the step-function shape might be an artifact of the KL divergence metric itself (or possibly due to softmax), rather than a true representation of activation plateaus. The logarithmic nature of KL divergence may amplify differences as they become larger, leading to a more pronounced blowup region and a flatter initial plateau region.

\begin{figure}[h]
  \centering
  \begin{subfigure}{0.497\textwidth}
  \includegraphics[width=\textwidth]{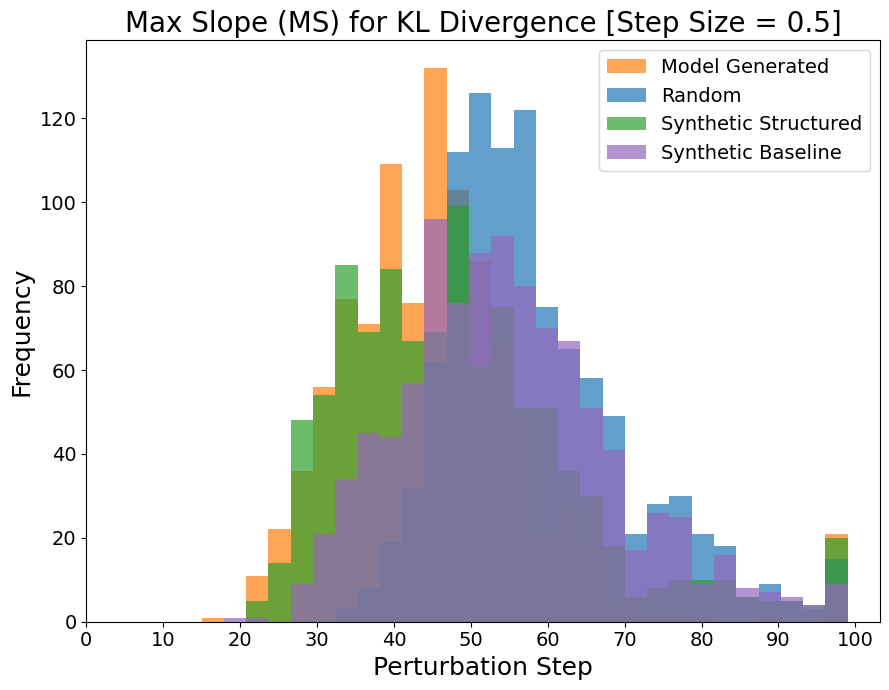}
  \end{subfigure}  
  \begin{subfigure}{0.497\textwidth}
  \includegraphics[width=\textwidth]{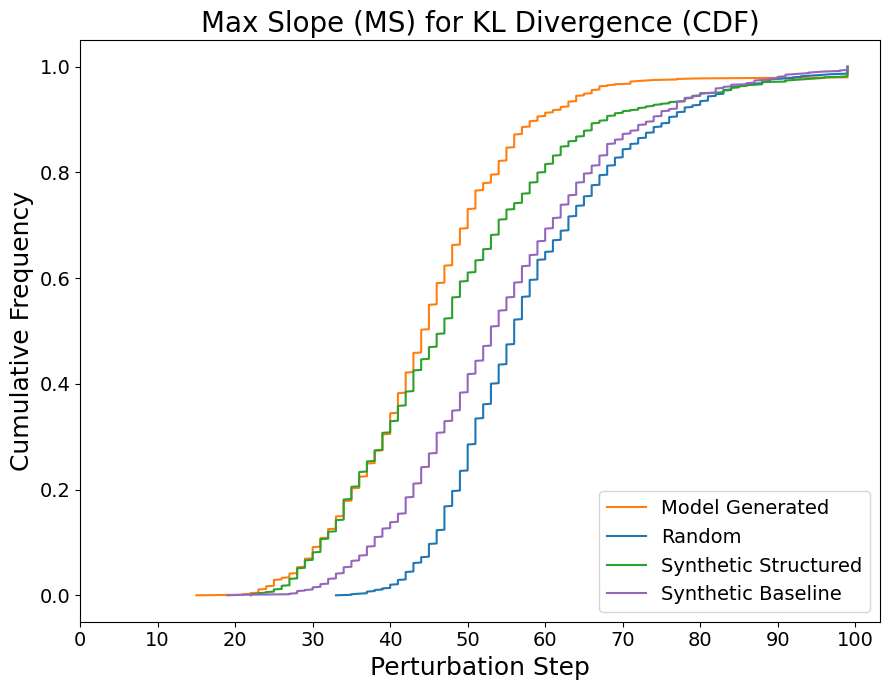}
  \end{subfigure}
  \begin{subfigure}{0.497\textwidth}
  \includegraphics[width=\textwidth]{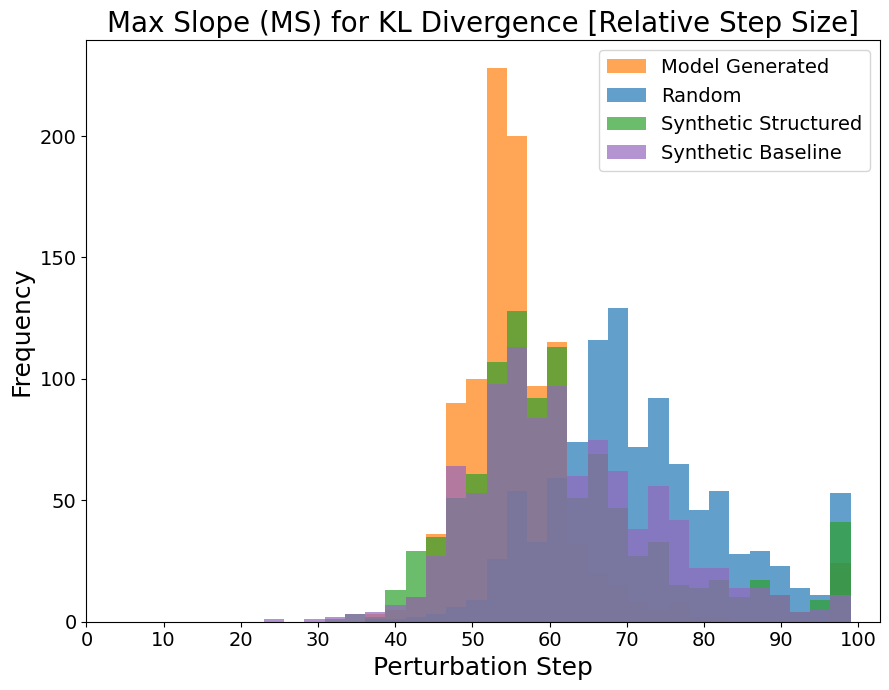}
  \end{subfigure}  
  \begin{subfigure}{0.497\textwidth}
  \includegraphics[width=\textwidth]{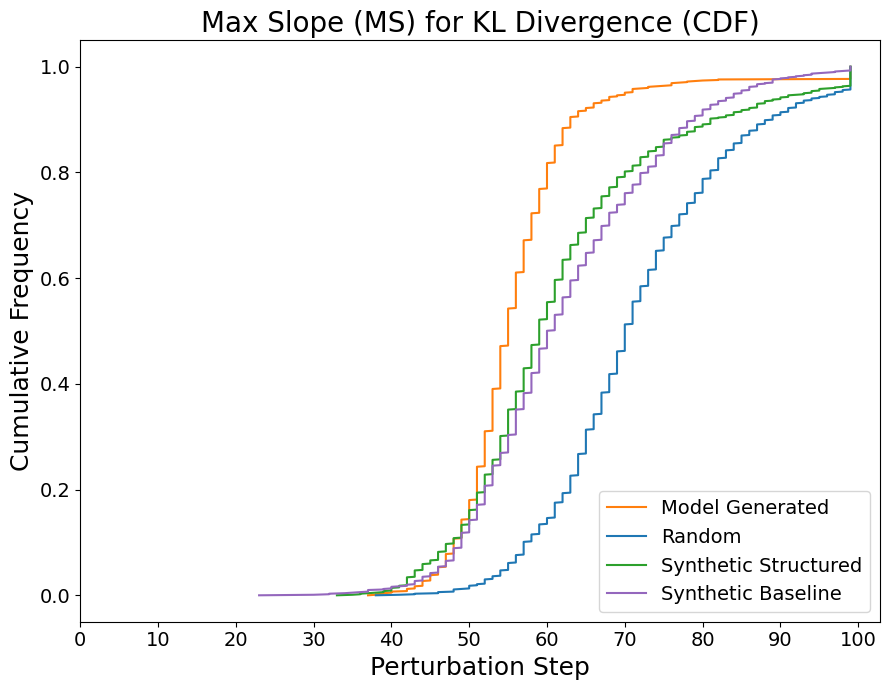}
  \end{subfigure}
  \caption{\footnotesize{The distributions of the MS steps for KL divergence of next-token prediction probabilities for perturbations with absolute step size (top) and relative step size (bottom) towards model-generated (orange), random (blue), synthetic-baseline (purple), and synthetic-structured (green) activations. The left column shows the counts of MS steps occurring in different bins along the length of the perturbation, and the right column shows the cumulative frequency for the same. We find that our results for KL divergence are similar to those for L2 distance.}}
  \label{figC1}
\end{figure}

With the mentioned caveats in mind, we perform perturbations at Layer 1 and observe their effect on KL divergence of the logits distribution instead of L2 distance at Layer 11. Figure \ref{figC1} illustrates the MS step distribution for KL divergence across different activation types. KL divergence blowups are more localized in the relative step size setup than L2 distance blowups, suggesting that the model’s output distribution is more robust to noise than the model’s final layer activations, only blowing up when more than $40\%$ of the base activation has been replaced.
Similar to the results for L2 distance, we find that perturbations towards synthetic-structured activations are more similar to perturbations towards model-generated activations than synthetic-baseline activations are. The difference between synthetic-structured and synthetic-baseline activations is more pronounced for KL divergence than L2 distance. 

\begin{table}[h]
  \centering
  \begin{tabular}{lccccccc}
    \multicolumn{7}{c}{Max Slope (MS) Step Distribution Statistics for KL divergence} \\
    \toprule
    \multirow{2}{*}{Activation Type} & \multicolumn{3}{c}{Absolute Step Size} & \multicolumn{3}{c}{Relative Step Size} \\
    \cmidrule(lr){2-4} \cmidrule(lr){5-7}
    & Mean & Std dev & KS & Mean & Std dev & KS \\
    \midrule
    Model Generated & 45.51 & 12.89 & 0.00 & 56.34 & 9.08 & 0.00 \\
    Random & 58.54 & \textbf{12.33} & 0.47 & 71.83 & \textbf{11.75} & 0.69 \\
    Synthetic Baseline & 54.79 & 14.02 & 0.32 & 62.41 & 12.05 & 0.32 \\
    Synthetic Structured & \textbf{48.79} & 15.64 & \textbf{0.13} & \textbf{61.86} & 13.51 & \textbf{0.26} \\
    \bottomrule
    \\
  \end{tabular}
  \caption{\footnotesize{We find that our results for KL divergence of next-token prediction probabilities are similar to those for L2 distance at Layer 11. This table contains the mean, standard deviation and KS statistic for MS step distributions for all the perturbations we perform. The KS statistic is measured against perturbations towards model-generated activations, with a lower value meaning higher similarity.}}
  \label{tabC1}
\end{table}

\newpage
\section{Isolating the effect of SAE reconstruction error} \label{app:recon}
\setcounter{figure}{0}
\setcounter{table}{0}
\renewcommand\thefigure{\Alph{section}.\arabic{figure}}
\renewcommand\thetable{\Alph{section}.\arabic{table}}

\begin{figure}[h]
  \centering
  \begin{subfigure}{0.497\textwidth}
  \includegraphics[width=\textwidth]{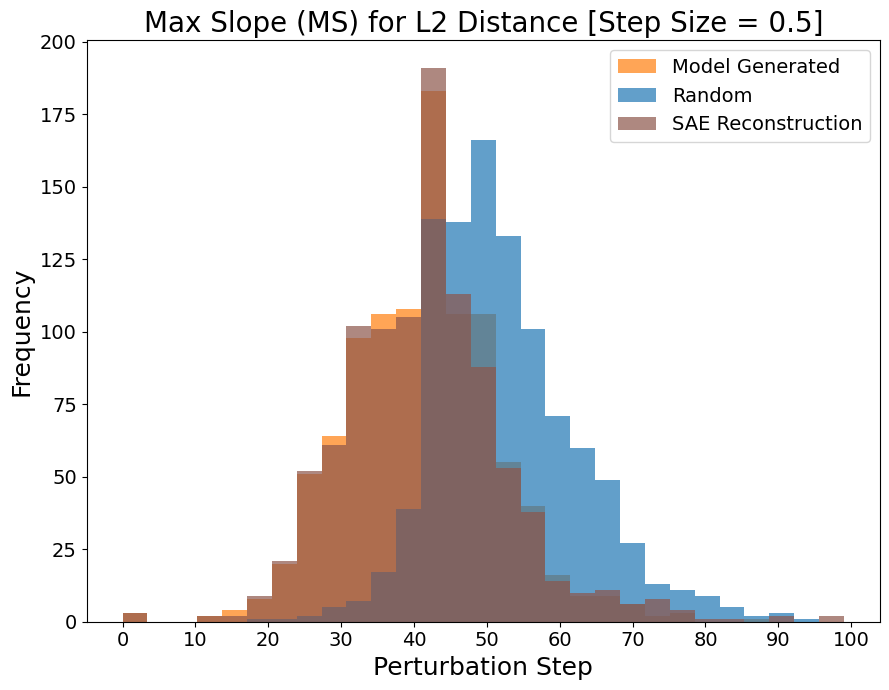}
  \end{subfigure}  
  \begin{subfigure}{0.497\textwidth}
  \includegraphics[width=\textwidth]{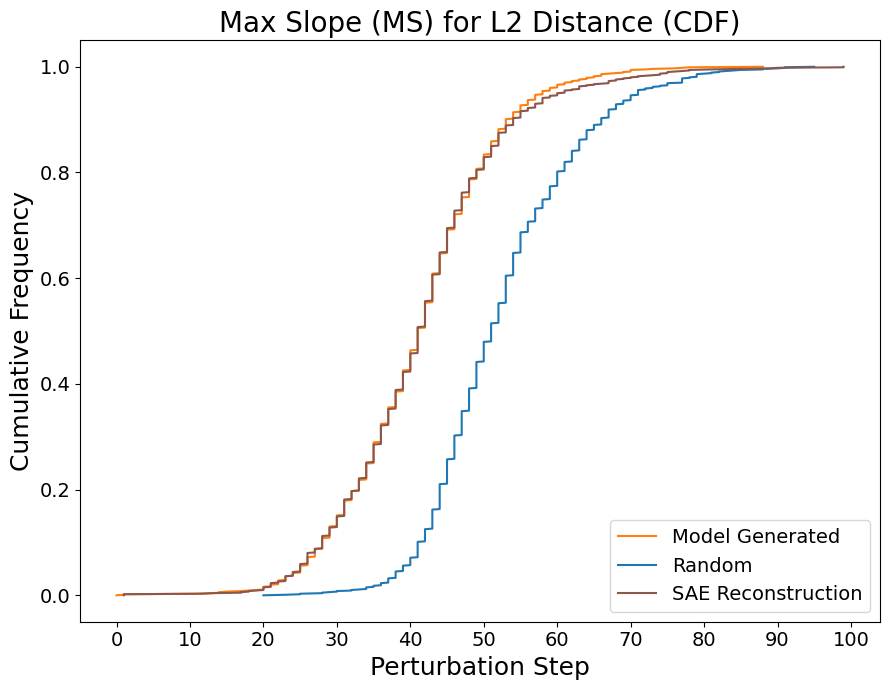}
  \end{subfigure}
  \begin{subfigure}{0.497\textwidth}
  \includegraphics[width=\textwidth]{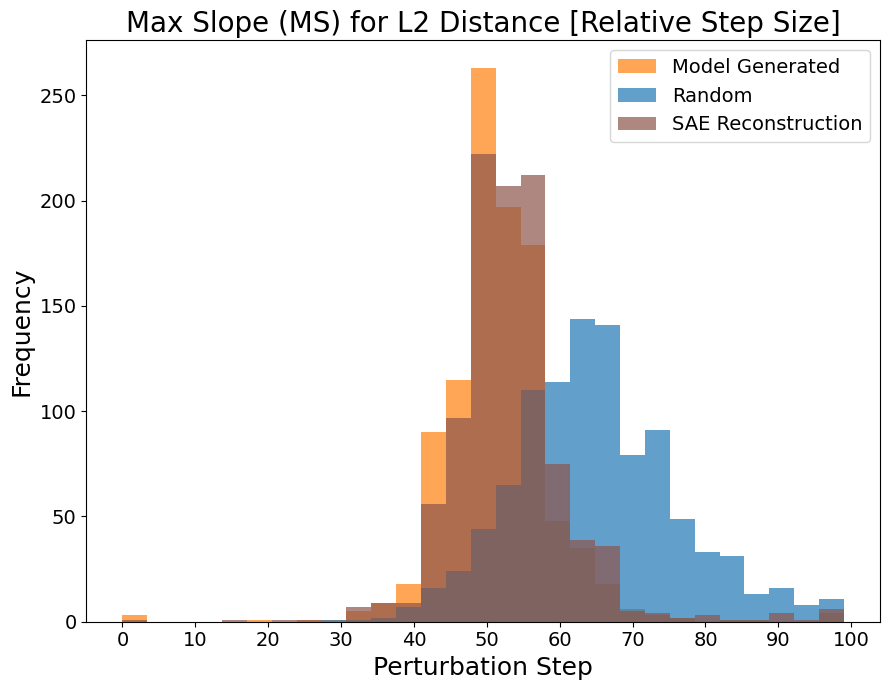}
  \end{subfigure}  
  \begin{subfigure}{0.497\textwidth}
  \includegraphics[width=\textwidth]{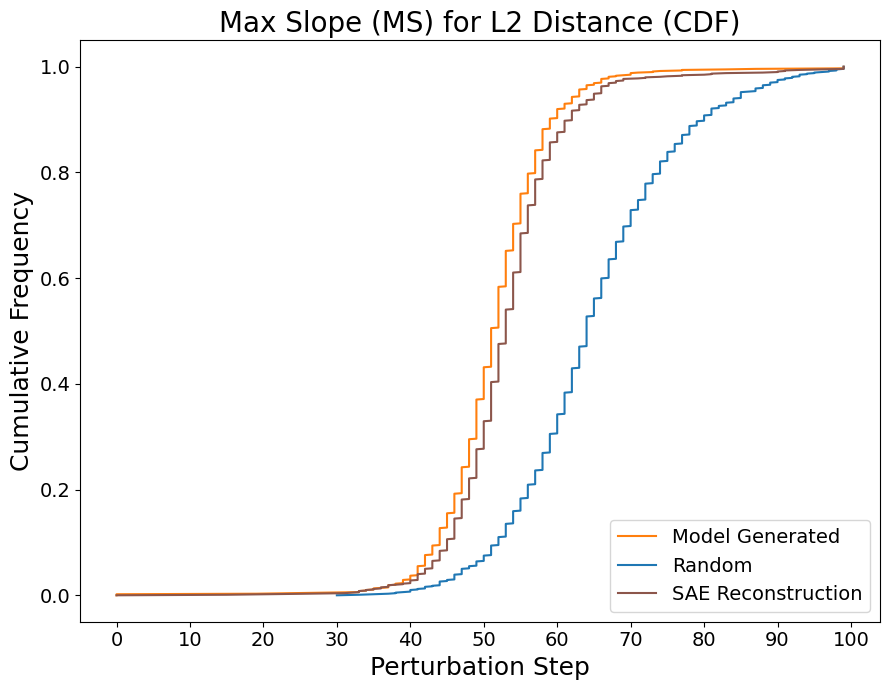}
  \end{subfigure}
  \caption{\footnotesize{The distributions of the MS steps for perturbations with absolute step size (top) and relative step size (bottom) towards random activations (blue), model-generated activations (orange), and their SAE reconstructions (brown). The left column shows the counts of MS steps occurring in different bins along the length of the perturbation, and the right column shows the cumulative frequency for the same. We find that perturbations towards model-generated activations and perturbations towards their SAE reconstructions are almost identical.}}
  \label{figD1}
\end{figure}

We denote the reconstruction of an activation A with \texttt{SAE(A) = decode(encode(A))}. To isolate the effect of SAE reconstruction error on the blowup location, we examine perturbations towards a reconstruction of a model-generated target activation \texttt{SAE(T)}. We compare these to perturbations towards model-generated activations and find that they are very similar, with blowups occurring slightly later for perturbations towards SAE reconstructions (Figure \ref{figD1}, Table \ref{tabD1}). We also find that reconstructions of model-generated activations also have plateaus around them. This shows that the majority of the difference in our synthetic activations comes from the heuristics we use to select latents, and not the SAE reconstruction error. 

This similarity suggests that SAE reconstructions behave like model-generated activations for the most part, and that the reconstruction error causes a small systematic shift in the blowup location. This points to some information loss that causes the model to respond slightly less to perturbations towards SAE reconstruction, which is relevant for interpreting experiments that use SAE latents.

\begin{table}[h]
  \centering
  \begin{tabular}{lccccccc}
    \multicolumn{7}{c}{Max Slope (MS) Step Distribution Statistics for SAE reconstructions} \\
    \toprule
    \multirow{2}{*}{Activation Type} & \multicolumn{3}{c}{Absolute Step Size} & \multicolumn{3}{c}{Relative Step Size} \\
    \cmidrule(lr){2-4} \cmidrule(lr){5-7}
    & Mean & Std dev & KS & Mean & Std dev & KS \\
    \midrule
    Model Generated & 41.11 & 10.40 & 0.00 & 51.60 & 7.82 & 0.00 \\
    Random & 52.49 & \textbf{10.21} & 0.45 & 65.01 & 11.19 & 0.61 \\
    SAE Reconstruction & \textbf{41.49} & 11.34 & \textbf{0.02} & \textbf{53.34} & \textbf{8.39} & \textbf{0.11} \\
    \bottomrule
    \\
  \end{tabular}
  \caption{\footnotesize{We find that perturbations towards model-generated activations are almost identical to perturbations towards their SAE reconstructions. This table contains the mean, standard deviation and KS statistic for MS step distributions for all the perturbations we perform. The KS statistic is measured against perturbations towards model-generated activations, with a lower value meaning higher similarity.}}
  \label{tabD1}
\end{table}

\newpage
\section{Properties of SAE latents in model activations} \label{app:stats}
\setcounter{figure}{0}
\setcounter{table}{0}
\renewcommand\thefigure{\Alph{section}.\arabic{figure}}
\renewcommand\thetable{\Alph{section}.\arabic{table}}

\begin{figure}[h]
  \centering
  \begin{subfigure}{0.475\textwidth}
  \includegraphics[width=\textwidth]{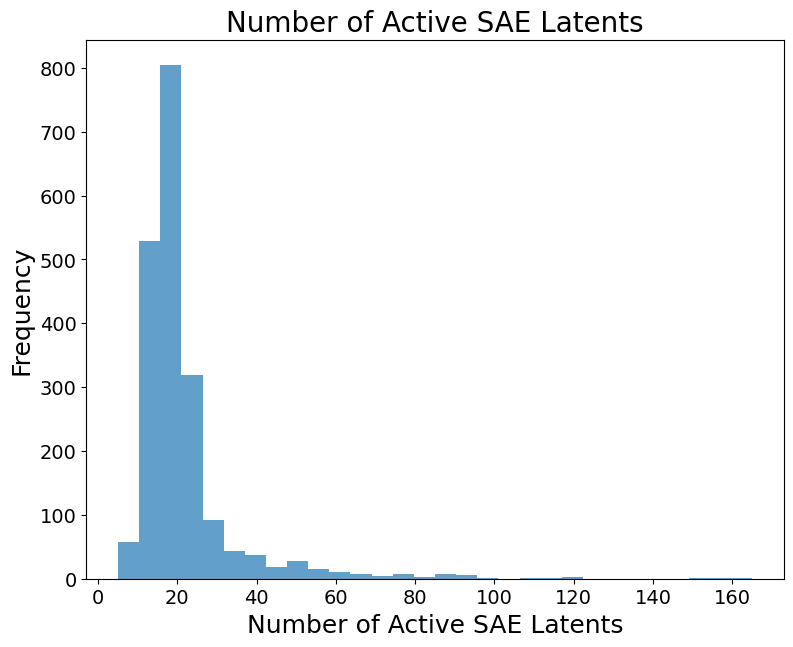}
  \end{subfigure}  
  \begin{subfigure}{0.497\textwidth}
  \includegraphics[width=\textwidth]{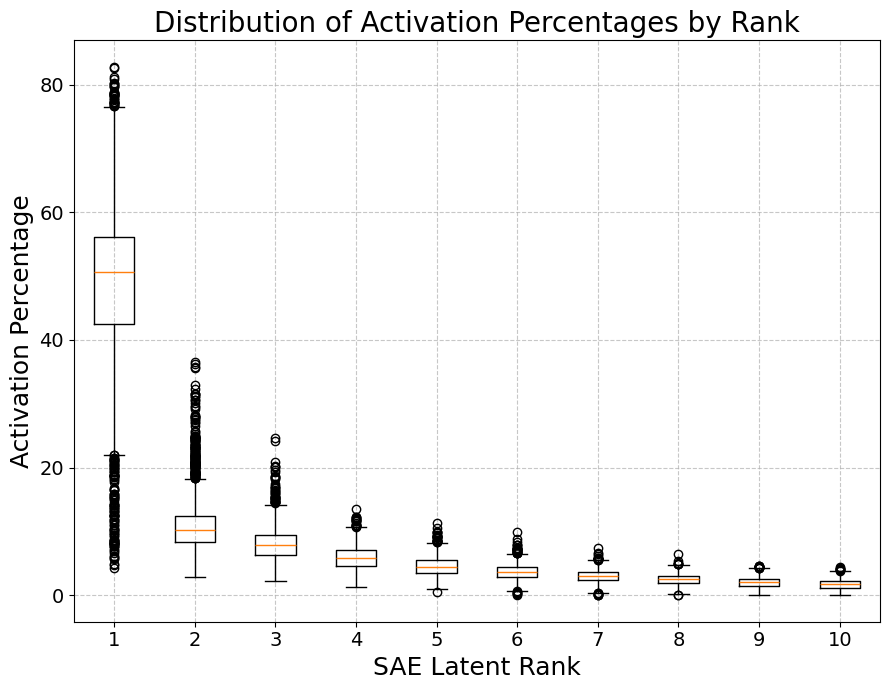}
  \end{subfigure}
  \caption{\footnotesize{The distribution of the total number of active SAE latents per activation (left) and the distribution of the percentage of the latent activation norm represented by the top $10$ active latents (right) aggregated over $2000$ activations.}}
  \label{figE1}

  \begin{subfigure}{0.476\textwidth}
  \includegraphics[width=\textwidth]{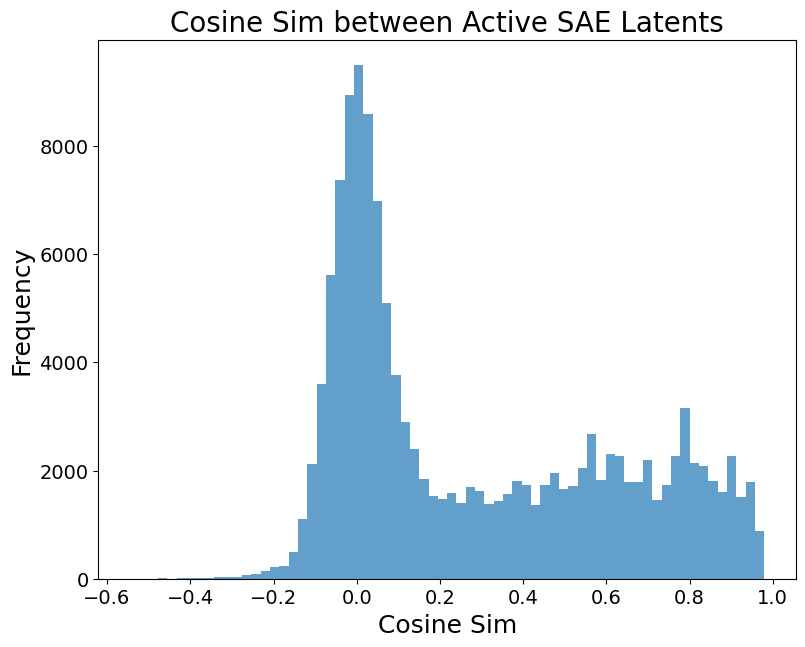}
  \end{subfigure}  
  \begin{subfigure}{0.497\textwidth}
  \includegraphics[width=\textwidth]{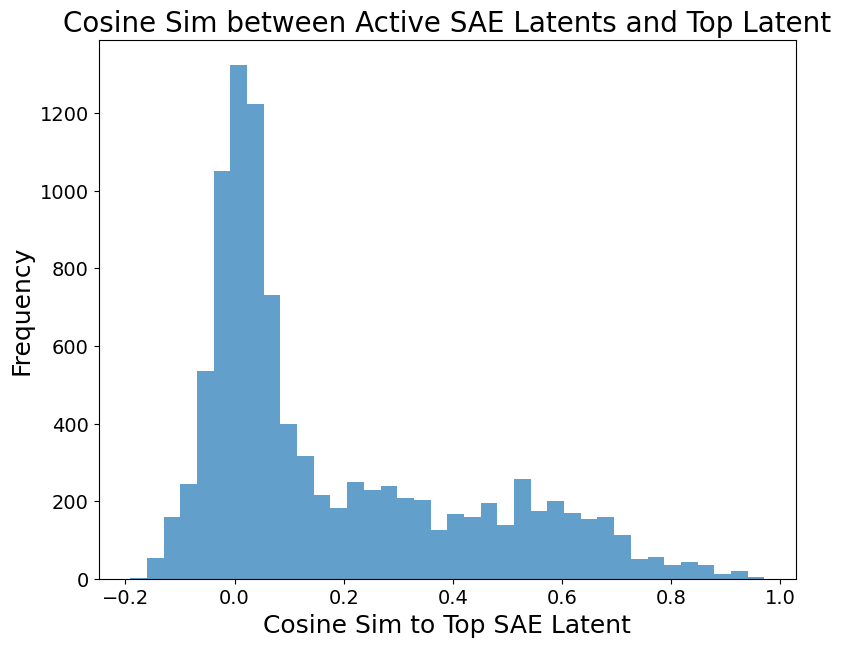}
  \end{subfigure}  
  \caption{\footnotesize{The distribution of cosine similarities between all active SAE latents per activation (left) and distribution of cosine similarities that active SAE latents have with the top SAE latent (right) aggregated over $2000$ activations.}}
  \label{figE2}
\end{figure}

We observe that model-generated activations with a low SAE reconstruction error contain approximately $21$ active SAE latents on average (Figure \ref{figE1} left). The distribution is narrow around the mean and falls off very rapidly. The top latent represents around $49\%$ of the total latent activation norm average (Figure \ref{figE1} right). The norm falls off rapidly thereafter, with the second top latent representing only around $10\%$ on average. The distribution flattens out afterwards where latter ranks have similar contribution to the norm.

Additionally, we find that model-generated activations are made up of SAE latents that have cosine similarity to one another of approximately $0.29$ on average (Figure \ref{figE2} left), with a distinct peak at $0$. SAE latents primarily have positive cosine similarity to the top SAE latent, with mean cosine similarity of $0.18$ (Figure \ref{figE2} right) and with a more pronounced peak at $0$.

\clearpage

\end{document}